\documentclass[conference]{IEEEtran}
\IEEEoverridecommandlockouts
\usepackage{cite}
\usepackage{amsmath,amssymb,amsfonts}
\usepackage[linesnumbered,ruled,vlined]{algorithm2e}
\usepackage{url}
\usepackage{makecell}
\usepackage{graphicx}
\usepackage{textcomp}
\usepackage{utfsym}
\usepackage{subcaption}
\usepackage{xcolor}
\usepackage{colortbl}
\usepackage{multirow}

\usepackage{bm}
\usepackage{amsmath}

\usepackage{listings}
\usepackage{listingsutf8}
\usepackage{xcolor}
\usepackage{tikz}

\usepackage{color}
\definecolor{pblue}{rgb}{0.13,0.13,1}
\definecolor{pgreen}{rgb}{0,0.5,0}
\definecolor{pred}{rgb}{0.9,0,0}
\definecolor{pgrey}{rgb}{0.46,0.45,0.48}
\definecolor{lblue}{rgb}{0.0,0.27,1.0}
\definecolor{lorange}{rgb}{0.99,0.44,0.0}
\definecolor{dgreen}{rgb}{0.23,0.49,0.14}

\usetikzlibrary{arrows,decorations.pathmorphing,decorations.text,backgrounds,fit,positioning,shapes.symbols,chains,calc,shadings,patterns,shapes,shapes.geometric,automata}
\tikzset{mynode/.style={draw=orange,ellipse,inner sep=1pt,font=\scriptsize,color=orange,anchor=south}}

\usepackage[colorlinks=true,linkcolor=blue,citecolor=blue]{hyperref} 

\usepackage{orcidlink}  

\hypersetup{
  colorlinks   = true,    
  urlcolor     = blue,    
  linkcolor    = red,    
  citecolor    = blue      
}

\makeatletter
\newenvironment{btHighlight}[1][]
{\begingroup\tikzset{bt@Highlight@par/.style={#1}}\begin{lrbox}{\@tempboxa}}
{\end{lrbox}\bt@HL@box[bt@Highlight@par]{\@tempboxa}\endgroup}

\newcommand\btHL[1][]{%
  \begin{btHighlight}[#1]\bgroup\aftergroup\bt@HL@endenv%
}
\def\bt@HL@endenv{%
  \end{btHighlight}%
  \egroup
}
\newcommand{\bt@HL@box}[2][]{%
  \tikz[#1]{%
    \pgfpathrectangle{\pgfpoint{1pt}{0pt}}{\pgfpoint{\wd #2}{\ht #2}}%
    \pgfusepath{use as bounding box}%
    \node[anchor=base west, fill=orange!30,outer sep=0pt,inner xsep=1pt, inner ysep=0pt, rounded corners=3pt, minimum height=\ht\strutbox+1pt,#1]{\raisebox{1pt}{\strut}\strut\usebox{#2}};
  }%
}
\makeatother

\lstdefinestyle{baseJava}{
    language=Java,
    basicstyle=\scriptsize,
    numbers=left,
    numberstyle=\scriptsize,
    stepnumber=1,
    numbersep=8pt,
    showstringspaces=false,
    breaklines=true,
    frame=single,
    showspaces=false,
    showtabs=false,
    moredelim=**[is][{\btHL[fill=gray!30,thin]}]{@BG@}{@BG@},
    commentstyle=\color{pgreen},
    keywordstyle=\color{pblue},
    stringstyle=\color{pred},
    rulecolor=\color{black},
    emph={
    Future,
    ChangeOrderResult,
    String,
    Boolean,
    ChangeOrderResult,
    boolean
    },
    emphstyle={\color{pblue}},
}

\lstdefinestyle{fig5}{
    language=Bash,
    basicstyle=\scriptsize,
    numbers=left,
    numberstyle=\scriptsize,
    stepnumber=1,
    numbersep=8pt,
    showstringspaces=false,
    breaklines=true,
    frame=single,
    showspaces=false,
    showtabs=false,
    keywordstyle=\color{pred},
    rulecolor=\color{black},
    moredelim=**[is][\color{pred}]{@}{@},
    emph={
    Normal,
    Run,
    Faulty
    },
    emphstyle=\textbf,
}

\usepackage[left=0.75in, right=0.75in, top=0.75in, bottom=0.75in]{geometry}

\def\BibTeX{{\rm B\kern-.05em{\sc i\kern-.025em b}\kern-.08em
    T\kern-.1667em\lower.7ex\hbox{E}\kern-.125emX}}
\begin{document}

\clearpage
\newgeometry{left=0.75in, right=0.75in, top=0.75in, bottom=0.77in}

\title{\vspace{4mm}\LARGE \bf ETac: A Lightweight and Efficient Tactile Simulation Framework for Learning Dexterous Manipulation \\

\vspace{-2mm}
\author{Zhe Xu\orcidlink{0009-0004-1043-4248}, Feiyu Zhao\orcidlink{0009-0006-6444-9896}, Xiyan Huang\orcidlink{0009-0007-5469-8298}, and Chenxi Xiao$^{*}$ \orcidlink{0000-0002-7819-9633}
\vspace{-2mm}

\thanks{This work was supported by the Natural Science Foundation of Shanghai under Grant 25ZR1402370. }

\thanks{The authors are with the School of Information Science and Technology at ShanghaiTech University, Shanghai 201210, China. \{xuzhe2023, zhaofy12024, huangxy22023, xiaochx\}@shanghaitech.edu.cn, $^*$Corresponding
author: Chenxi Xiao.}
\thanks{Project webpage: \href{https://lassford.github.io/ETac/}{https://lassford.github.io/ETac/}.}

}

}

\maketitle

\begin{abstract}
Tactile sensors are increasingly integrated into dexterous robotic manipulators to enhance contact perception. 
However, learning manipulation policies that rely on tactile sensing remains challenging, primarily due to the trade-off between fidelity and computational cost of soft-body simulations. 
To address this, we present ETac, a tactile simulation framework that models elastomeric soft-body interactions with both high fidelity and efficiency.
ETac employs a lightweight data-driven deformation propagation model to capture soft-body contact dynamics, achieving high simulation quality and boosting efficiency that enables large-scale policy training. 
When serving as the simulation backend, ETac produces surface deformation estimates comparable to FEM and demonstrates applicability for modeling real tactile sensors.
Then, we showcase its capability in training a blind grasping policy that leverages large-area tactile feedback to manipulate diverse objects. 
Running on a single RTX 4090 GPU, ETac supports reinforcement learning across 4,096 parallel environments, achieving a total throughput of 869 FPS. 
The resulting policy reaches an average success rate of 84.45\% across four object types, underscoring ETac's potential to make tactile-based skill learning both efficient and scalable.

\end{abstract}

\section{Introduction}\label{sec:intro}

Robots rely on tactile sensing to perceive and interpret physical interactions in contact-rich environments. Following this trend, tactile sensors have been widely integrated into robotic hardware, including grippers~\cite{murali2018learning, guo2024proprioceptive}, dexterous hands~\cite{ganguly2020graspingindark, zhao2024embedding}, arms~\cite{si2023robotsweater, crowder2025social}, and feet~\cite{zhang2021foot}. 
Alongside advances in hardware, robots have demonstrated improved dexterous manipulation skills, such as grasping~\cite{murali2018learning, ganguly2020graspingindark} and object reorientation~\cite{yin2024learning}. 
Collectively, these developments not only enhance the fidelity and safety of physical interactions, but also broaden the range of real-world applications where robots can be effectively deployed~\cite{tegin2005tactile, yamaguchi2019recent}.

Building on these advances, there is growing interest in developing tactile-based skills for robotic manipulation.
A widely adopted approach is reinforcement learning (RL), which has shown strong potential for acquiring diverse manipulation skills.
Since RL requires extensive interaction data collected through repeated task executions, simulation environments are often preferred for their low cost, high repeatability, and minimal risk to hardware~\cite{wang2024lessons, lee2024dextouch}.
However, learning tactile-based skills through RL remains challenging, primarily due to the lack of tactile simulators that are both high-fidelity and computationally efficient. 

\begin{figure}[t]
    \centering
    \includegraphics[trim=0pt 0pt 0pt 0pt, clip, width=\linewidth]{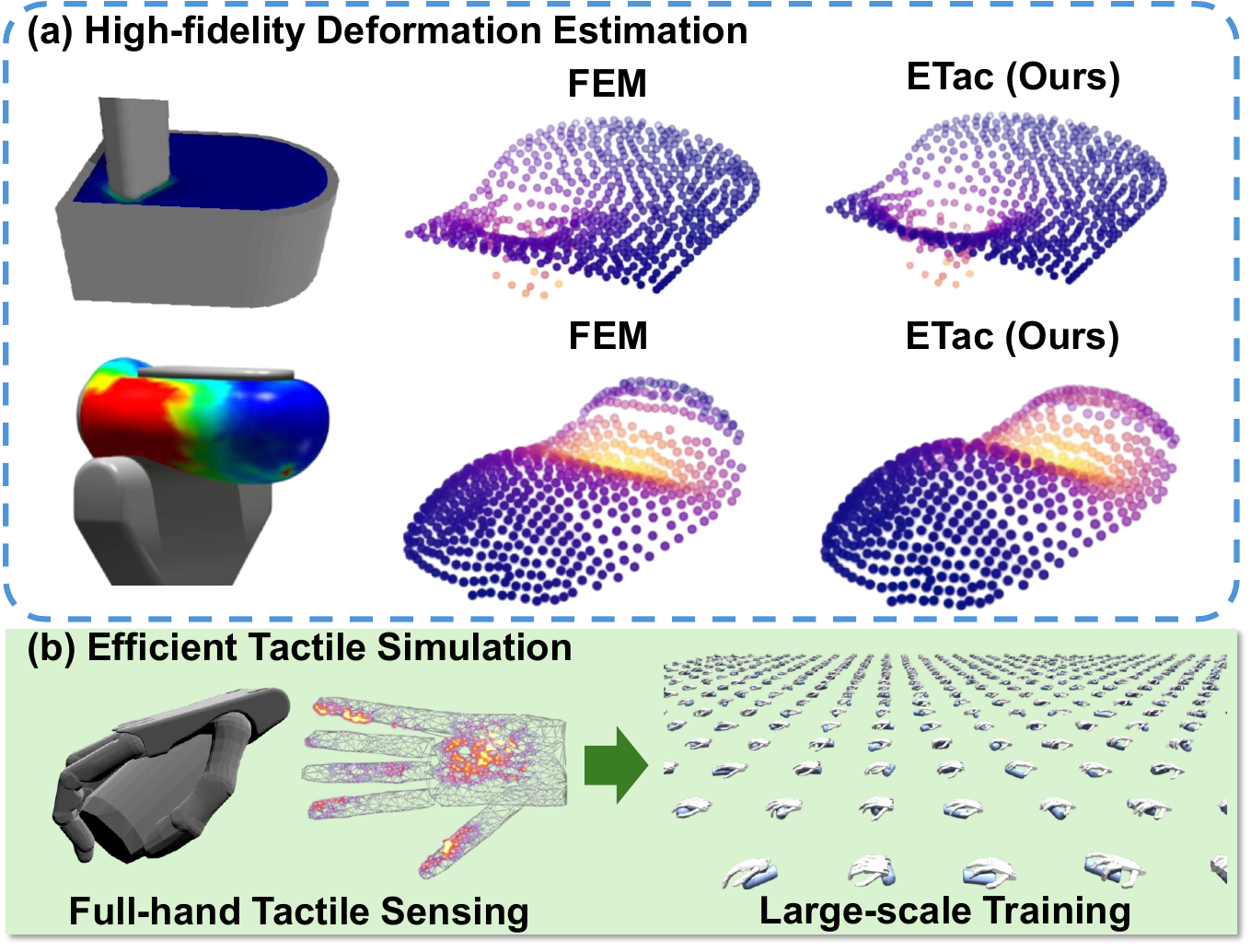}       \caption{\textbf{ETac: a lightweight and efficient tactile simulation framework.} (a) ETac estimates elastomer surface deformations with fidelity comparable to that of FEM. (b) It enables large-area tactile sensing for dexterous manipulators while supporting large-scale RL training.}
    \label{fig:teaser}
\end{figure}

Currently, a common choice for state-of-the-art tactile simulators is based on soft-body physics engines. 
For instance, simulators built on the Finite Element Method (FEM) can capture high-quality contact deformations during physical interactions~\cite{lan2022affine, si2024difftactile}. 
However, many of these simulators are developed as standalone tools, their integration with multi-fingered robotic hands remains limited and largely inaccessible to the broader research community. 
On the other hand, the intensive data exchange among numerous mesh elements makes FEM-based models several orders of magnitude slower than rigid-body physics engines~\cite{xu2023efficient}. 
Such computational costs not only slows RL training but also prevents large-scale parallel rollouts, which are essential for stable gradient estimation for policy updates~\cite{rudin2022learning}.
To mitigate these issues, some studies have explored accelerating simulation with simplified models~\cite{wang2022tacto, si2022taxim, yin2024learning, akinola2024tacsl}. 
Nevertheless, these approaches typically tend to emphasize local surface deformations at contact areas. They fail to sufficiently capture strain propagation, which is crucial for accurately estimating global elastomer deformation. 

In this work, we introduce \textbf{ETac}, a tactile simulation framework that models the soft-body behavior of tactile sensors with both high computational efficiency and simulation fidelity. Compared to previous tactile simulation approaches~\cite{si2022taxim, akinola2024tacsl, yin2024learning}, the key innovation of ETac is its learnable deformation propagation model, which integrates an analytical model with a compact residual network. This hybrid approach enables ETac to capture the nonlinear dynamics arising from anisotropy and surface curvature, producing high-quality simulations on curved surfaces, such as those of dexterous robotic hands. Furthermore, ETac features a lightweight design that supports large-scale parallelization, achieving performance comparable to or exceeding that of existing tactile simulators.

Our experiments demonstrate that ETac not only achieves high-fidelity sensor simulation but also provides new possibilities for robot learning. 
First, when used as the simulation backend of tactile sensors, it faithfully reproduces surface deformation patterns upon contact. These deformation patterns can be further mapped to real sensors' output signals, enabling the real-to-sim modeling for tactile sensor hardware. 
Second, ETac’s efficiency overcomes the scalability bottleneck in RL training for contact-rich tasks. 
In a challenging dexterous blind grasping task within Isaac Gym~\cite{makoviychuk2021isaac}, ETac accomplished RL training across 4,096 parallel environments at up to a total of 869 FPS on a single RTX 4090 GPU. 
This represents an 11× increase in FPS and 128× parallelization capacity compared to an FEM solver under the same GPU hardware. 
Furthermore, the policy trained with ETac’s tactile feedback achieved an 84.45\% success rate (a 21.48\% improvement over a non-tactile baseline).
These results highlight the potential of ETac as a scalable and versatile platform for tactile-driven skill learning in dexterous robots, contributing to the embodied AI community.

In summary, the contributions of this paper are as follows:

\begin{itemize} 
    \item \textbf{ETac:} A lightweight and efficient tactile simulation framework for learning dexterous manipulation.
    \item A learnable deformation propagation model that estimates the elastic response of deformable tactile sensor surfaces.
    \item Comprehensive evaluations of ETac in terms of simulation fidelity, computational efficiency, and effectiveness in training RL policies.
\end{itemize}

\section{Related Work}\label{sec:rela}

\subsection{Tactile Sensing and Manipulation}
Tactile sensing has become a fundamental component of robotic hands.
Over the past decades, tactile sensors have been integrated into various manipulators such as the Svelte Hand~\cite{Zhao2023GelSightSH}, Eyesight Hand~\cite{Romero2024EyeSightHD}, and F-Tac~\cite{zhao2024embedding}, progressively achieving higher resolution and broader coverage.
With these advances, researchers have been developing diverse tactile-based skills, including in-hand manipulation~\cite{yin2023rotating, yuan2024robot}, grasping~\cite{murali2018learning}, and object placement~\cite{chen2023sequentialdexteritychainingdexterous}, to mention a few.

Despite this progress, a critical gap is the lack of effective approaches to fully exploit the information that modern tactile sensors conveyed.
Traditional methods often rely on simplified tactile representations, encoding feedback as binary contact states or low-dimensional force/torque signals~\cite{ganguly2020graspingindark, yin2024learning}.
These representations fail to capture the rich, high-fidelity information provided by modern tactile sensors.
Reinforcement learning has emerged as a promising alternative, offering a data-driven framework that can adapt to diverse sensor modalities and leverage high-dimensional tactile feedback for skill acquisition~\cite{xu2023efficient, lee2024dextouch, wang2024lessons}.
Since data collection from real-world interactions is prohibitively expensive, most reinforcement learning approaches still depend heavily on simulation.
Nevertheless, accurately reproducing tactile feedback in simulation remains a challenge~\cite{xu2023efficient, si2024difftactile}.

\subsection{Simulating Tactile Sensors}

Simulators of tactile sensors aim to obtain simulated response signals in contact events. 
Developing such simulators is technically challenging, mainly due to the high degrees of freedom in soft-body models, and the complex non-linear contact dynamics. 
High-fidelity methods, including finite element models (FEM)~\cite{lan2022affine, si2024difftactile, li2025taccel} and the Material Point Method (MPM)~\cite{hu2018moving, hu2019chainqueen}, have demonstrated realistic simulation results.
However, the considerable computational cost limits their scalability in large-scale RL training.

To address efficiency bottlenecks, several studies adopt approximation strategies, as summarized in Tab.~\ref{tab:related}. 
A common approach is to approximate deformation via penetration distance~\cite{wang2022tacto, yin2024learning, akinola2024tacsl}, which offers efficiency but fails to simulate strain propagation to surrounding regions, thereby compromising realism. Taxim~\cite{si2022taxim} introduces an affine linear deformation propagation model that partially alleviates this issue, yet it overlooks higher-dimensional effects such as anisotropy and surface curvature. 
To overcome these limitations, ETac incorporates a lightweight neural network-based propagation model that simulates the non-linear influences between elements on deformation, thereby achieving high fidelity while preserving the computational efficiency necessary for scalable RL training.

\section{Methodology}\label{sec:meth}

\begin{figure*}[t]
    \centering
    \includegraphics[trim=0pt 0pt 0pt 0pt, clip, width=\textwidth]{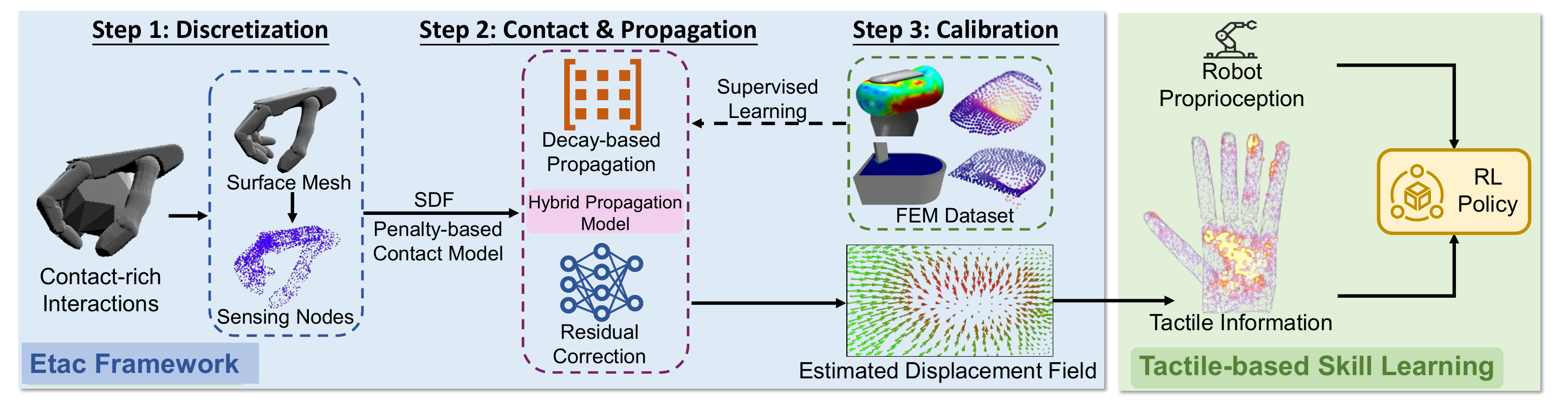}  \caption{\textbf{Overview of the ETac pipeline.} The manipulator’s surface mesh is discretized into nodes, whose contact dynamics are learned by a hybrid deformation propagation model. We calibrate the propagation model using data from FEM simulation to ensure fidelity. 
    The computed displacement field of the nodes serves as the tactile input to the RL policy during skill learning.}
    \label{fig:pipeline}
\end{figure*}

Our research aims to develop tactile simulators that achieve both high simulation accuracy and computational efficiency.
To this end, we propose ETac, a lightweight tactile simulation framework, with its response learned to be consistent with high-fidelity, computationally expensive FEM simulations (Sec.\ref{sec:etac}). 
Furthermore, to substantiate ETac’s effectiveness in supporting scalable reinforcement learning for contact-rich manipulation, we establish a tactile-based blind grasping benchmark addressed through RL (Sec.~\ref{sec:blindgrasp}). The system pipeline is illustrated in Fig.~\ref{fig:pipeline}.

\begin{table}[t]
\centering
\caption{\textbf{Comparison of lightweight tactile simulation methods.} Element Dimension refers to whether displacement is a scalar (1 dim) or vector (3 dims), while Propagation and Type indicate whether and how nodes influence each other.}

\label{tab:related}
\resizebox{\columnwidth}{!}{%
\begin{tabular}{ccccc}
\hline
\textbf{Method}    & \textbf{Element Dimension}   & \textbf{Propagation}  &  \textbf{Propagation Types} & \textbf{Differentiability} \\ \hline
Tacto~\cite{wang2022tacto}               &        1    &   -  & -  & - \\
Yin et al. ~\cite{yin2024learning}                               &  1           & -    & -  & -              \\
TacSL~\cite{akinola2024tacsl}                    &  3               & -  & -  & \checkmark               \\
Taxim~\cite{si2022taxim}                        &  3         & \checkmark    & Linear & -      \\ \hline
\textbf{ETac (Ours)}          &  3 & \checkmark    & Linear \& Nonlinear & \checkmark      \\ \hline
\end{tabular}%
}
\end{table}

\subsection{Lightweight Tactile Simulation Framework}
\label{sec:etac}
\textbf{Overview.} 
To efficiently simulate the physical response of tactile sensors with high fidelity, \textbf{ETac} employs a particle-based simulation technique that avoids costly global information updates on dense grids (as required in FEM).
The method first discretizes the manipulator’s surface mesh into a set of nodes in 3D space. These nodes detect contact with external objects, and then determine elastomer deformation by propagating contact-induced displacements through node dynamics.

To implement this, contact events are first detected using signed distance fields (SDFs)~\cite{KaolinLibrary} between nodes and object meshes, including both external object contact and self-contact of the hand. Here, we refer to nodes in direct contact with objects as \textit{active nodes}. Active node's contact dynamics have been extensively studied in previous works.
Here we follow~\cite{xu2023efficient, akinola2024tacsl} by adapting the penalty-based contact model. 
However, such model only models local displacements at the contact location. 
Due to the elasticity of the material, displacements at active nodes should also propagate through the medium of the elastomer and influence the neighboring region.
This propagation effect is critical for simulating soft sensor behavior realistically. 
Yet, recent lightweight tactile simulators either neglect it entirely~\cite{wang2022tacto, akinola2024tacsl, yin2024learning} or use overly simplified linear models~\cite{si2022taxim}, which fail to capture such propagation dynamics with high fidelity.

\textbf{Modeling Contact Propagation Dynamics.} 
To address the aforementioned limitations, we propose a deformation propagation model. 
This propagation model computes the displacements of nodes that are not in direct contact with external objects (denoted as \textit{passive nodes}) based on the displacements of active nodes. Here, we denote active node's displacements as:  $\bm{U}_a = \left[\bm{\hat{u}}_1, \bm{\hat{u}}_2, ..., \bm{\hat{u}}_m\right]$.

Then, the displacements of passive nodes are inferred jointly using two components: 
(1) a decay-based linear displacement propagation, and 
(2) a residual correction that incorporates nonlinear effects. 
The decay-based model provides an empirical prior for propagation, yielding smooth, translation-invariant predictions based on relative indentation locations. 
The residual component then captures high-dimensional effects that are difficult to model explicitly and provides additive corrections via a network. 
Together, these two components complement each other by balancing the generalization of physical priors and the expressiveness of the data-driven network, 
and thus improving simulation accuracy (as evaluated in Sec.~\ref{sec: exp1}).

1) \textit{Decay-based Linear Propagation}. 
Surface displacements propagated to passive nodes decay with distances from the indented location.
Following prior work~\cite{vinogradova2003interaction}, we use an exponential kernel to model this spatial decay. 
Specifically, for an active node $p_j$, its propagated displacement to a passive node $p_i$ is computed as:
\begin{equation}
\bm{u}_i^j = e^{-\alpha || \bm{p}_j - \bm{p}_i||_2} \cdot  \bm{\hat{u}}_j
\label{Eq:linear_decay}
\end{equation}
where $\bm{u}_i^j$ denotes the propagated displacement, $\bm{p}_j$ and $\bm{p}_i$ are the undeformed positions of $p_j$ and $p_i$, respectively, and $\alpha$ is a learnable parameter that reflects the material-specific attenuation of deformation. 
To enable efficient paralleled propagation calculation, the overall effect between $m$ active nodes and $(n - m)$ passive nodes is modeled using a propagation matrix $\bm{\Omega} \in \mathbb{R}^{(n - m) \times m}$, where $\bm{\Omega}[i, j] = e^{-\alpha \lVert \bm{p}_j - \bm{p}_i \rVert_2}$. The propagated displacements for all passive nodes are then computed as $\bm{\Omega} \bm{U}_a$.

Note this spatial decay-based model alone provides only an approximation of the propagation process.
It cannot fully capture non-linear effects caused by factors such as surface curvature and anisotropy. 
This is not only because linear model's limited representability, but also because there is a subtle error: {$\bm{U}_a$ are not solely caused by direct contact; they already include mutual interactions among active nodes. Hence, directly applying the decay-based propagation ``double-counts'' these influences. }
Unlike Taxim~\cite{si2022taxim}, which adopts a computationally expensive matrix inversion to explicitly correct the inter-node interactions, our approach learns to correct for this and other residual errors jointly using a neural network in the next stage.

\begin{figure*}[t]
    \centering
    \includegraphics[trim=0pt 0pt 0pt 0pt, clip, width=\textwidth]{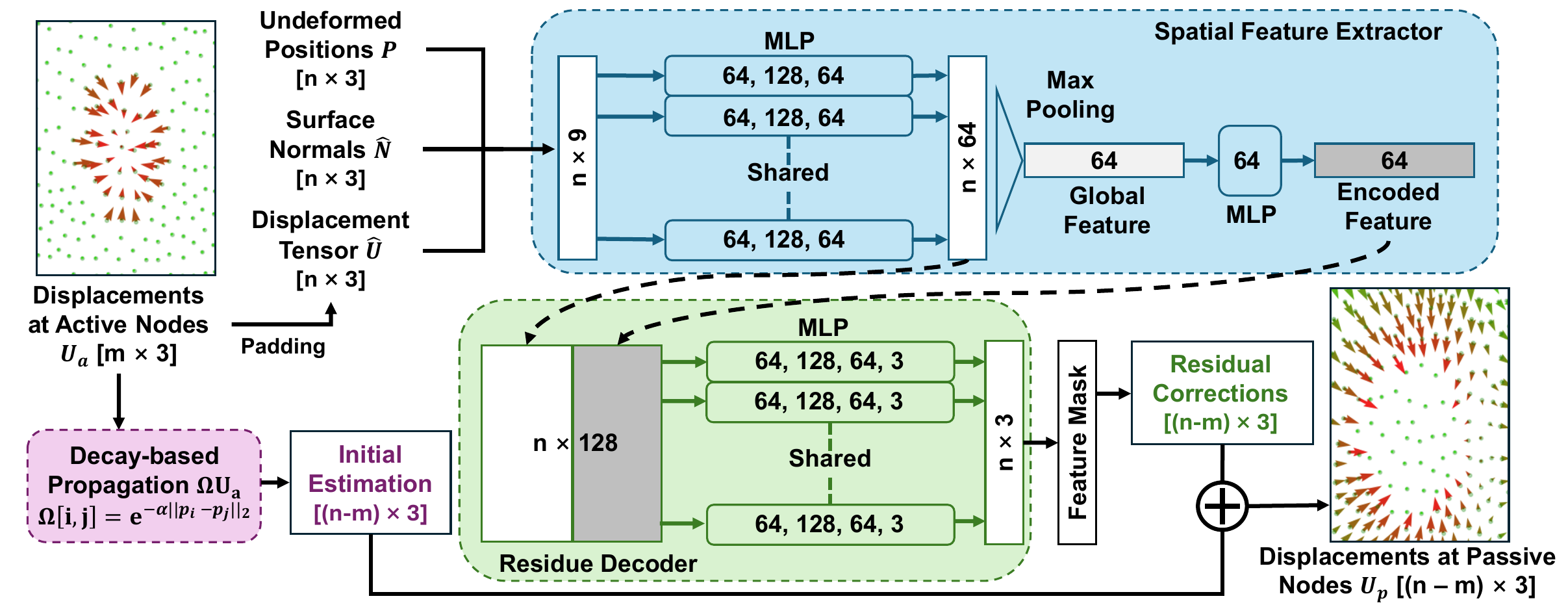 }  \caption{\textbf{Computational process of the propagation model}. Displacements of passive nodes (nodes not in direct contact) are estimated using a decay-based model combined with a residual correction network.}
    \label{fig:propagation}
\end{figure*}

2) \textit{Residual Displacement Correction. }
{We address the above issues and other residual effects by introducing a residual network $\Phi$, as illustrated in Fig.~\ref{fig:propagation}.
It uses a lightweight PointNet~\cite{qi2017pointnet} encoder to extract features from elastomer's surface, followed by an MLP-based decoder that estimates node-wise residual displacements.

Specifically, the input comprises the undeformed positions $\bm{P} \in \mathbb{R}^{n \times 3}$ and normals $\bm{N} \in \mathbb{R}^{n \times 3}$ of all nodes, together with the displacements $\bm{\hat{U}} \in \mathbb{R}^{n \times 3}$ (the displacements of passive nodes are padded with zeros), forming an $n \times 9$ dimensional feature tensor. 
The encoder extracts point-wise features using shared MLPs and aggregates them via Max-Pooling into a 64-dimensional global feature vector. This global feature is further encoded by another MLP, then broadcast and concatenated with per-node features to form an $n \times 128$ tensor. 
Finally, the shared MLP decoder produces the node-wise $n \times 3$ residual displacements for all nodes,
{following which a binary mask $M_p \in \{0, 1\}^n$ is used to retrieve residual displacements of passive nodes, denoted as
$\Phi_p(\bm{P}, \bm{\hat{U}}, \bm{N}) \in \mathbb{R}^{(n-m) \times 3}$}. }
Together with the decay-based propagation, the final estimated displacements of all passive nodes $\bm{U}_p$ are calculated as:
\begin{equation}
\bm{U}_p = \bm{\Omega}  \bm{{U}}_a + \Phi_p(\bm{P}, \bm{\hat{U}}, \bm{N})
\label{eq:deformation}
\end{equation}

\textbf{Learn Propagation Dynamics From FEM. }
The kernel parameter $\alpha$ in Eq.~(\ref{Eq:linear_decay}) and the network weights jointly determine the characteristics of deformation propagation in Eq.~(\ref{eq:deformation}). 
As these parameters implicitly encode the material properties, we calibrate them in advance using an FEM-based simulation. 
Specifically, we construct an elastomer model in a finite element simulator (Isaac Gym~\cite{makoviychuk2021isaac} in our case, though other simulators are also applicable), with Young’s modulus and Poisson’s ratio matched to those of real-world materials.
We conduct loading experiments with varying indentation locations, depths, and indenter orientations to generate diverse data.
At each time step, we record the index of active nodes and the displacement tensor of all surface nodes. 
The collected data are then used to minimize the following loss function via gradient descent:

\begin{equation}
    Loss=\underset{\alpha, \Phi_p}{\ min} ||\bm{U}_p - \bm{U}_p^{*}||_2 
\end{equation}
where $\bm{U}_p^{*}$ denotes oracle displacements of passive nodes from FEM simulation.

\subsection{RL Training: Tactile-Based Dexterous Blind Grasping}
\label{sec:blindgrasp}
To demonstrate ETac’s capability in enabling scalable reinforcement learning, we develop a tactile-based dexterous blind grasping task. 
Here, we used ETac as the tactile simulation's backend. 
The resultant displacement tensor from ETac is provided as input to the manipulation policy. This state variable design follows protocols in ~\cite{narang2021sim, amri2024transferring}.

The objective of the task is to manipulate and grasp a variety of objects using a ShadowHand manipulator~\cite{shadowrobot} in a vision-deprived setting, thereby emphasizing the usage of simulated tactile feedback. 
The manipulated objects include both geometric primitives and common household items such as an egg, a soda can, and a cube. 
Each grasp trial begins with initial contact between the hand and the object, with both the object's pose and the contact location randomized. As the object may not be centered in the palm, the hand must gradually conform its fingers to the object's surface and ultimately lift it.

We train the policy using Proximal Policy Optimization (PPO)~\cite{schulman2017proximal}. The input state to the policy is defined as $S_t = (s_t^r, I_t, a_{t-1})$, where \( s_t^r \) represents robot joint's proprioceptive states, \( I_t \) denotes tactile observations (i.e., the displacement tensor described above), and \( a_{t-1} \) is the previous action. Given this input, the policy outputs a six DoF wrist translation and rotation in Euler angles (floating base controlled using PD controllers), along with 18 joint torque values applied to the actuators. The reward function is defined below:
\begin{equation}
    r = r_{\text{proximity}} + r_{\text{lift}} + r_{\text{goal}} + r_{\text{contact}}
\end{equation}
where the proximity reward \( r_{\text{proximity}} \) encourages minimizing the distance between the hand and the object. 
The contact reward \( r_{\text{contact}} \) promotes broader contact areas to enhance grasp stability. 
Once a grasp is established, the lift reward \( r_{\text{lift}} \) is activated to encourage object lifting, following the design in~\cite{xu2023unidexgrasp}. 
Finally, the goal reward \( r_{\text{goal}} \) provides a bonus for successful grasping and penalizes failures. Details of the reward design are provided in Appendix section.

\begin{figure*}[t]
    \centering
    \includegraphics[trim=0pt 5pt 0pt 5pt, clip, width=\textwidth]{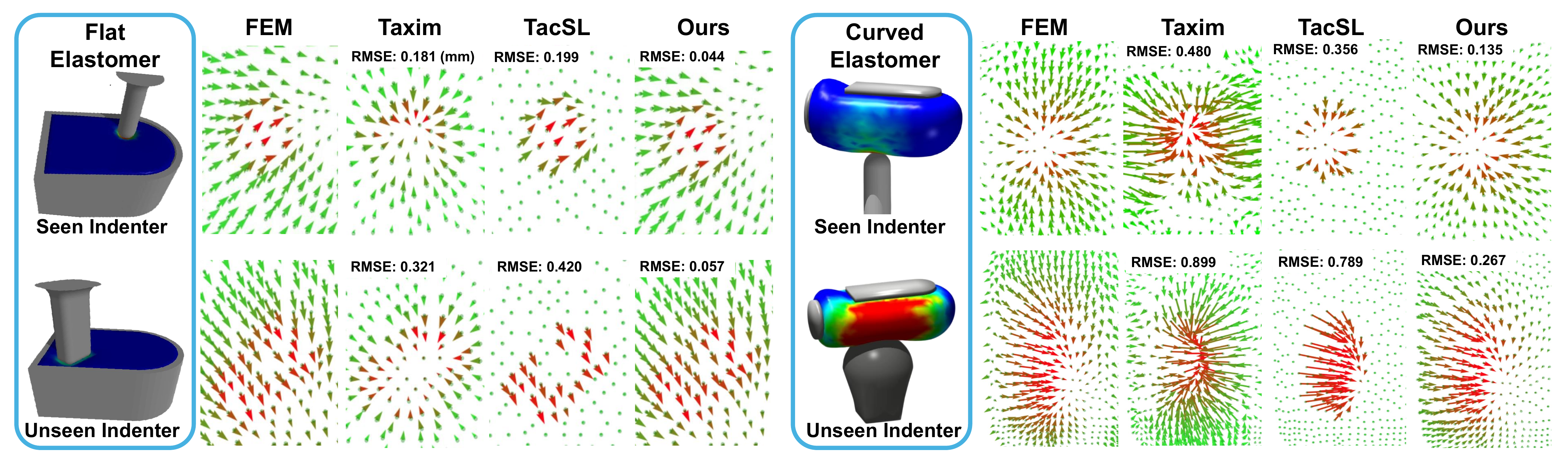}
    \caption{\textbf{Comparison of elastomer surface displacement fields estimated by different methods.} Arrows show node-wise displacement direction and magnitude, with color shifting to red as $z$-axis displacement increases.}

    \label{fig:exp1}
\end{figure*}

\section{Experiments}
\label{sec:result}

In this section, we evaluate ETac through experiments to address two questions: 
{\color{orange}(\textbf{Q1})} Does it provide high-quality tactile information? and 
{\color{magenta}(\textbf{Q2})} Is it efficient for parallel RL training of dexterous manipulation policies?

\subsection{Q1: Simulation Fidelity} \label{sec: exp1}

\textbf{Deformation Estimation. }We first evaluate ETac’s simulation quality by estimating sensor elastomer deformation. 
Indentation experiments are conducted on two elastomer types: a flat elastomer~\cite{huang2025twintac} and a curved BioTac-style elastomer~\cite{narang2021sim}. The experiments were with eight indenters, each with over 12,000 frames of data. 
Loading positions cover the full elastomer surface with depths from 0 to 4 mm.

\begin{table}[t]
\centering
\caption{The average RMSE (mm) between estimated displacements and the FEM simulation results.}
\label{tab:qualitative_quality}
{%
\begin{tabular}{lll}
\hline
               & \textbf{Flat Elastomer} & \textbf{Curved Elastomer} \\ \hline \hline
TacSL~\cite{akinola2024tacsl} &   0.194 $\pm$ 0.076   &    0.445 $\pm$ 0.079     \\
Taxim~\cite{si2022taxim}          &   0.163 $\pm$ 0.048   &    0.447 $\pm$ 0.358    \\
\textbf{Ours} (Linear Decay Only)          &   0.151 $\pm$ 0.058   &    0.256 $\pm$ 0.062    \\
\textbf{Ours} (Residual Network Only)            &   0.074 $\pm$ 0.032   &   0.128 $\pm$ 0.055   \\  
\textbf{Ours} (Full Pipeline)           &   \textbf{0.058 $\pm$ 0.034}   &    \textbf{0.116 $\pm$ 0.049}    \\ \hline
\end{tabular}%
}
\end{table}

Using this dataset, we compare ETac with two particle-based simulators: Taxim \cite{si2022taxim} and TacSL \cite{akinola2024tacsl}. 
The quality metric is based on displacements' node-wise RMSE against FEM ground truth (Tab.~\ref{tab:qualitative_quality}). 
On the flat elastomer, ETac achieves 0.058 mm RMSE, outperforming TacSL (0.194 mm) and Taxim (0.163 mm). 
The advantage is larger on the curved elastomer surface, with ETac at 0.116 mm versus 0.445 mm (TacSL) and 0.447 mm (Taxim). Fig.~\ref{fig:exp1} shows qualitative results from four randomly selected cases. 
In all examples, ETac most closely matches the FEM results and achieves the lowest RMSE. 
By contrast, Taxim \cite{si2022taxim} performs well on flat surfaces but struggles with curved deformations, while TacSL \cite{akinola2024tacsl} produces noticeably sparser outputs due to the absence of a deformation propagation mechanism.

Furthermore, we conduct an ablation study to evaluate the necessity of decomposing ETac's propagation model into decay-based propagation and residual correction components. For this evaluation, we retrained two ablated baselines.
Using only decay-based propagation significantly degrades performance, with RMSE increasing to 0.151 mm on the flat elastomer and 0.256 mm on the curved elastomer.
Relying solely on the residual network structure yields better results (0.074 mm and 0.128 mm, respectively) but still falls short of the full model with better physical prior knowledge for modeling attenuation.
Their combination achieves the best performance in deformation estimation.

\begin{figure}[t]
    \centering
    \includegraphics[trim=0pt 0pt 0pt 0pt, clip, width=\linewidth]{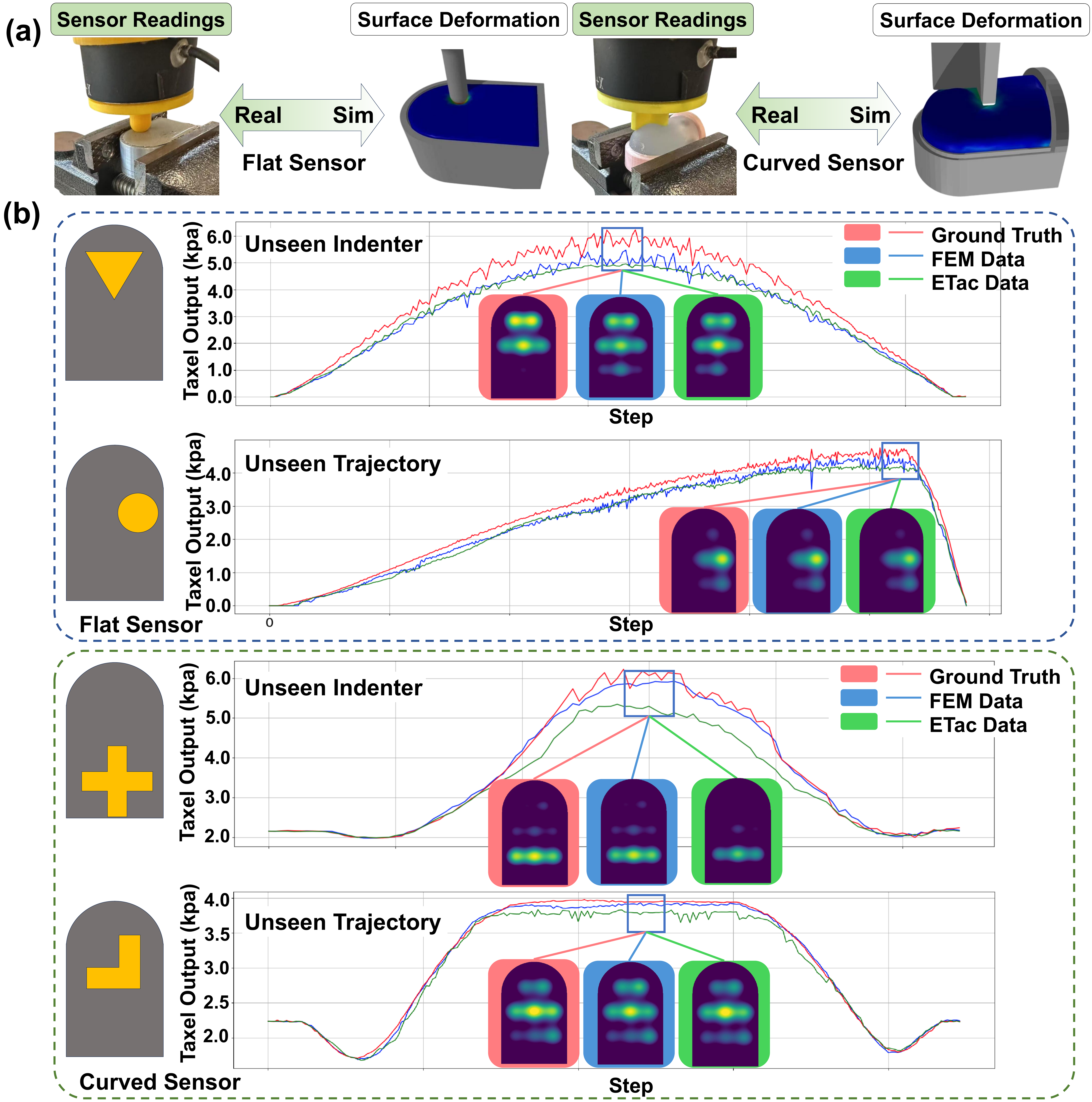} 
    \caption{\textbf{Predicting signals of real sensors. }(a) Paired data collection in real and simulated settings. (b) Comparison of sensor output predictions using ETac and FEM data. ``Unseen Trajectory'' denotes novel loading patterns with seen indenters.}
    \label{fig:sim2real}
\end{figure}

\begin{figure*}[t]
    \centering
    \includegraphics[trim=0pt 10pt 0pt 10pt, clip, width=\textwidth]{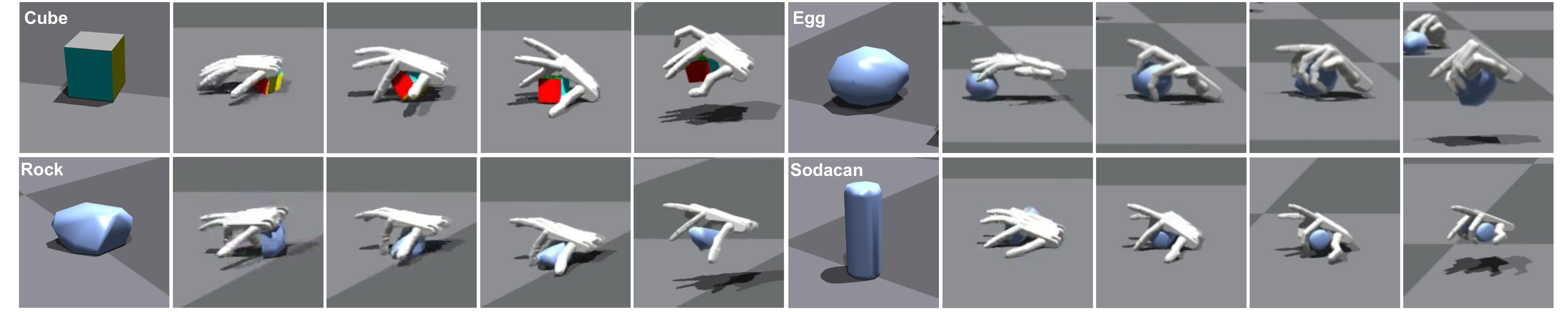}
    \caption{\textbf{Demonstrations of Blind Grasping Skill.} The process of successfully grasping 4 different objects over time.}
    \label{fig:process}
\end{figure*}

\begin{table*}[h]
\centering
\caption{Success rates (\%) of blind grasping policies under different settings. }
\label{tab:my-table}
{%
\resizebox{\textwidth}{!}{%
\begin{tabular}{cccccc}
\hline
            & {Cube}              & {Egg}               & {Rock}              & {Sodacan}           & {Average}           \\ \hline \hline
{Baseline (Object Pose)} & $63.79 \pm 36.25$          & $72.05 \pm 30.22$          & $47.66 \pm 43.63$          & $68.70 \pm 12.12$          & $62.97 \pm 37.82$          \\
{Fingertip Sensors}      & $74.35 \pm 5.24$           & $86.46 \pm 10.57$          & $70.36 \pm 15.30$          & $60.43 \pm 3.01$           & $72.90 \pm 21.06$          \\
{Full-hand Sensors}      & {\bf 86.95 $\pm$ 4.65}     & {\bf 92.98 $\pm$ 3.91}     & {\bf 78.42 $\pm$ 9.03}     & {\bf 82.45 $\pm$ 11.03}    & {\bf 84.45 $\pm$ 13.09}    \\ \hline
\end{tabular}%
}
}
\label{tab:res1}
\end{table*}

\textbf{Real-to-Sim Tactile Sensor Modeling. }Moreover, we validated ETac’s ability to simulate real sensors by replicating the experimental setup from \cite{huang2025twintac}. 
The goal is to reproduce the outputs of the tactile sensors consisting of eight MEMS-based pressure-sensing taxels. The evaluation is conducted on two types of elastomers: one flat and one curved.
While the original study \cite{huang2025twintac} used Isaac Gym’s FEM backend for Real2Sim, we employed ETac for benchmarking. 
We collected paired data by subjecting both real sensors and their simulated counterparts to identical indentation conditions with four different indenters (Fig.~\ref{fig:sim2real}(a)). 
During this process, we recorded signals from the real sensor’s eight taxels and corresponding simulated surface deformations from ETac and FEM. 
This dataset was then used to train a PointNet-based model for predicting real sensor responses (refer to Appendix). 
Qualitative results are shown in Fig.~\ref{fig:sim2real}(b), including the time series response of the most heavily loaded taxel and the outputs of all eight taxels at peak load. 
Our analysis shows that the model trained using ETac data achieved L1 losses of 3.94\% and 3.61\% for the flat and curved sensors, respectively, comparable to the FEM backend (2.46\% and 2.75\%). 
These findings demonstrate that ETac is a viable alternative to FEM for data-driven sim-to-real learning.

Based on all the experimental results, we conclude that ETac provides high-quality information for tactile simulation {\color{orange}(\textbf{Q1})}.

\subsection{Q2: Evaluation on RL-based Blind Grasping}\label{sec: exp:grasp}

We next evaluate ETac’s effectiveness for training reinforcement learning policies in a vision-deprived grasping task (refer to Sec.~\ref{sec:blindgrasp}). 
We tested three sensor settings:

\begin{enumerate} 
\item \textbf{Full-Hand Sensors}: {The tactile sensor covers the front side of the hand, demonstrating the simulator’s capability to handle a large number of nodes (1,000 nodes). }
\item \textbf{Fingertip Sensors}: Tactile sensors cover robot’s fingertips (150 nodes). This follows conventional Shadow Hand which has 5 BioTac fingertip sensors \cite{shadowrobot}. 
\item \textbf{Baseline}: As a comparison baseline, the hand is not equipped with any tactile sensors. Instead, the oracle 6D pose of the object is provided as input to the policy. This setting follows the configuration used in \cite{xu2023unidexgrasp}. \end{enumerate}

\begin{figure}[t]
    \centering
    \includegraphics[trim=00pt 00pt 0pt 00pt, clip, width=\linewidth]{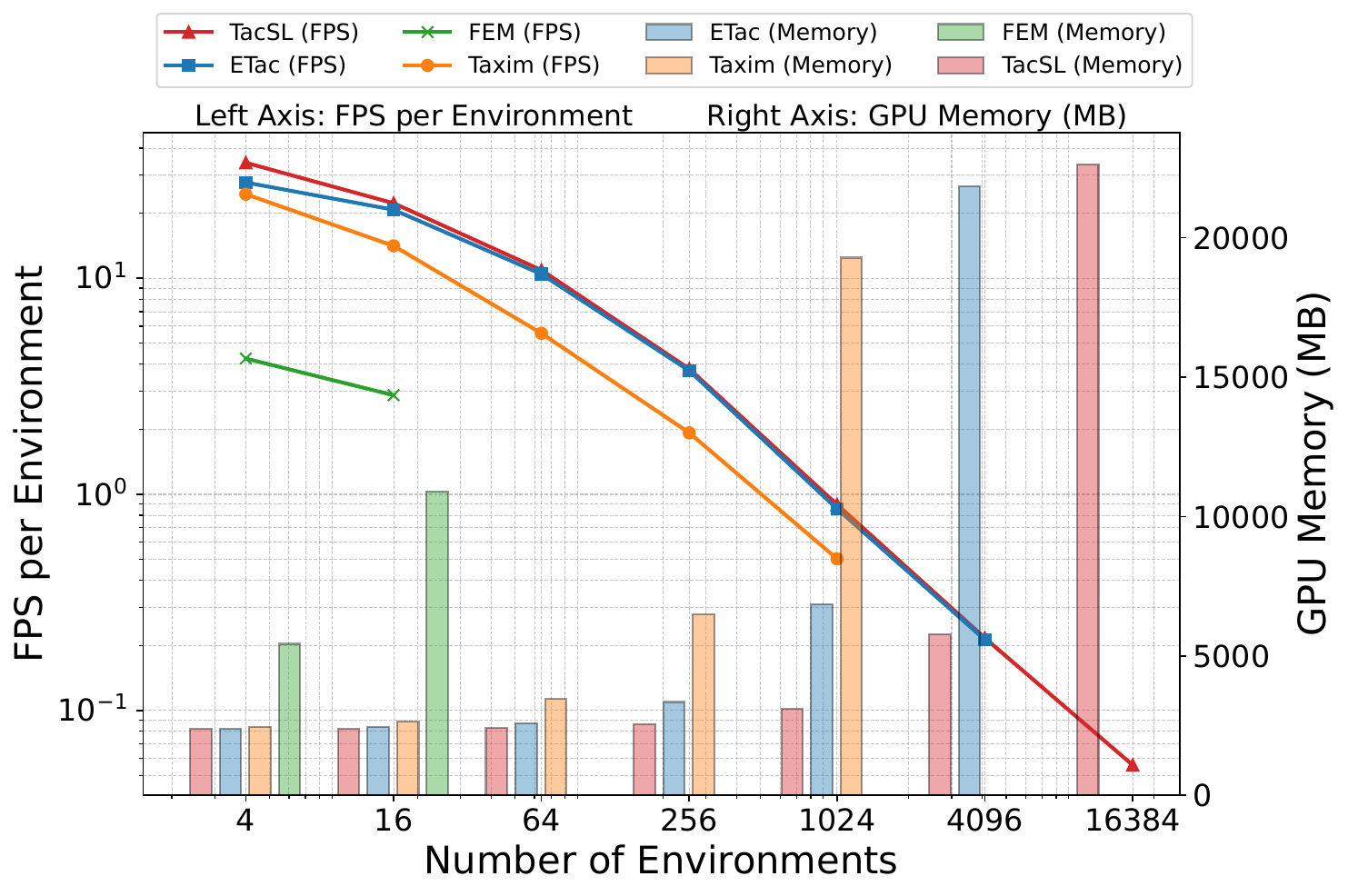}
    \caption{Comparison of RL training performance. }
    \label{fig:effciency}
\end{figure}

\textbf{Computational Efficiency.} A key factor for ETac’s usability in parallel RL training is its computational efficiency. 
We evaluated this in the blind grasping task, comparing ETac with Taxim~\cite{si2022taxim}, TacSL~\cite{akinola2024tacsl}, and an FEM solver (Fig.~\ref{fig:effciency}). 
When simulating a full-hand sensory setup on an RTX 4090 desktop GPU (24~GB RAM), the FEM solver can handle only up to 32 environments in parallel, whereas ETac efficiently scales to support up to 4096 environments. At 32 environments, ETac maintains 14.85 FPS per environment, which is faster than FEM (2.38 FPS per env). 
For total FPS achieved, ETac scales efficiently with more environments: reaching 669, 956, 878, and 869 total FPS at 64, 256, 1024, and 4096 environments, respectively. 
This outperforms Taxim (508, 620, 650 FPS at 64, 256, 1024 environments, but out-of-memory when scale up to 4096 envs) and is comparable to TacSL (698, 975, 918, 886 FPS at 64, 256, 1024, 4096 envs).  
ETac’s memory usage is lower than Taxim’s due to less model parameters, but higher than TacSL’s because of its propagation calculation. 
This trade-off enables higher-fidelity deformation estimation, making ETac suitable for tactile-based dexterous manipulation.

\textbf{Grasping Performance.} Our tactile simulation enables the successful acquisition of blind grasping skills. The learned skill performances are shown in Tab.~\ref{tab:res1}. Specifically, the \textit{Full-Hand Sensor} setup achieved the highest performance, with an average grasp success rate of 84.45\% across four objects. Fig.~\ref{fig:process} illustrates successful grasping processes under this configuration, where the hand continuously adjusted its pose based on tactile feedback, gradually enclosed and ultimately lifted the object. This demonstrates that ETac effectively provides rich contact information for perception, which is crucial for successful training. 
When limited to \textit{Fingertip Sensors}, the average success rate decreased to 72.90\%. Nonetheless, both tactile configurations outperformed the baseline without tactile sensing, which relied solely on the object’s pose and could not reflect physical characteristics such as size and shape. 
These results confirm that ETac provides effective tactile information that support parallel RL training for acquiring tactile dexterous manipulation policies {\color{magenta}(\textbf{Q2})}.

\section{Conclusions}\label{sec:con}
In this paper, we introduced ETac, a novel framework for efficient and high-fidelity tactile sensor simulation. 
By distilling high-fidelity FEM contact dynamics to a lightweight data-driven deformation propagation model, ETac accurately estimates elastomer surface deformations, achieving performance comparable to FEM and superior to baselines. 
ETac supports manipulation skill learning in two ways. First, it enables the real-to-sim modeling for tactile sensors by functioning as a simulation backend. Second, its low computational cost enables large-scale parallel RL training.
We demonstrated its utility in a blind grasping task, where ETac achieved 864 FPS when simulating full-hand tactile sensing across 4,096 environments. 
Using ETac, we successfully trained policies that leverage simulated full-hand tactile feedback to lift objects, highlighting its potential for contact-rich manipulation.  

{While ETac demonstrates significant advantages in efficiency and simulation fidelity, we acknowledge several limitations.
First, regarding scalability in calibration, ETac relies on high-fidelity FEM simulations to calibrate the propagation model. Extending it to new sensor geometries or materials currently involves creating specific oracle models and re-calibration, imposing a pre-computation overhead.
Second, concerning application scope, this work focuses on validating the framework's capacity for large-scale parallel training. Although sensor-level real-to-sim alignment is validated, full sim-to-real transfer of the learned blind grasping policy remains our ongoing work.
Finally, in terms of interaction dynamics, ETac is currently integrated with a rigid-body physics backend. Although it accurately estimates the deformation of sensor elastomers, the two-way coupling dynamics (specifically the influences onto object states) is simplified, limiting support for deformable objects and soft–soft contact.
These limitations outline the improvement directions for future work.}

\appendix
\subsection{Reward Design for Training Blind Grasping Policy}
\label{sec: reward}

Our reward design used for blind grasping policy (Sec.~\ref{sec:blindgrasp}) comprises four sub-goals. Here we give a detailed description of each term (note that all the $\omega_{**}$ here denote weight hyper-parameters):

\textbf{Proximity Reward. }First, we encourage hand to be positioned within close proximity to the object: 
\begin{equation}
r_{proximity} = -\omega_{ho} \cdot D_{ho} -\omega_{fo} \cdot D_{fo}    
\end{equation}
where $D_{ho}$ is the distance between the object and the origin of the hand, $D_{fo}$ denotes the summed distances from the object to all fingertips.

\textbf{Lifting Reward. }When the following proximity conditions are satisfied: $D_{ho} \leq \lambda_{l1}$ and $D_{fo} \leq \lambda_{l2}$  ($\lambda_{l1}$ and $\lambda_{l2}$ are user-defined thresholds), the lifting behavior is activated. In this case, the agent continuously receives a lifting reward:
\begin{equation}
    r_{lift} = \omega_{lift} \cdot \left( f_z - 2 \cdot \Delta h \right)
\end{equation}
where $f_{z}$ is the force exerted on the hand root along the lifting direction (z-axis), and $\Delta h$ denotes the difference between the target height and the current object height. This follows \cite{xu2023unidexgrasp}.

\textbf{Contact Reward. }To prevent slippage, it is beneficial to enlarge the contact area between the hand and the object. To encourage this, we define the contact reward $r_{contact}$ as:
\begin{equation}
    r_{contact} = \text{nonzero}(S_{cr})/\text{len}(S_{cr}) + \begin{aligned}
    \omega_a \cdot
 \begin{cases}
    R_{a} & \text{if $n_{a} > 0$}\\
   -0.1   & \text{otherwise}
 \end{cases}
 \end{aligned}
\end{equation}
where $n_{a}$ denotes the number of active nodes (i.e., nodes in direct contact), and $R_{a} = n_a / n$ represents their proportion relative to the total number of nodes $n$. The one-hot vector $S_{cr}$ encodes the contact between the object and different linkages of the hand (e.g., the palm, the proximal, middle, and distal of each finger). The term $\text{nonzero}(S_{cr})/\text{len}(S_{cr})$ aims to distribute the contact points across multiple fingers.

\textbf{Goal Reward. }Finally, a goal reward is assigned based on whether hand has successfully grasped the object:
\begin{equation}
    r_{goal} = \begin{aligned}
 \begin{cases}
   \omega_{goal} & \text{if success}\\
   -2\cdot \omega_{goal}  & \text{if fail} \\
   0   & \text{otherwise}
 \end{cases}
 \end{aligned}
\end{equation}
the success condition is: $\Delta h \leq \lambda_{goal}, n_{a} > 0$, with $\lambda_{goal}$ as a user-defined threshold. 
A failure occurs if the object is not lifted within the allotted time or if it falls after being lifted.

In our implementation, we set $\omega_{ho} = 1.0$, $\omega_{fo} = 0.8$, $\omega_{lift} = 1.5$, $\omega_a = 5.0$, $\omega_{goal} = 1.0$, and $\lambda_{l1} = 0.12$, $\lambda_{l2}=0.25$ and $\lambda_{goal}=0.03$, respectively. We trained the policy with PPO \cite{schulman2017proximal} (Sec.~\ref{sec: exp:grasp}) using 1024 parallel environments. The PPO hyperparameters are shown in Tab.~\ref{tab:ppo}. 

\begin{table}[h]
\centering
\caption{PPO hyperparameters for blind grasping policy.}
\label{tab:ppo}
\resizebox{0.95\columnwidth}{!}{
\begin{tabular}{ll}
\hline
Hyperparameters                  & Value                \\ \hline \hline
Hidden Layers in Actor Network  & [512, 512, 256, 128] \\
Hidden Layers in Critic Network & [512, 512, 256, 128] \\
Activation                      & elu                  \\
Initial Learning Rate                   & 5e-4                 \\
Learning Rate Schedule                  & adaptive                 \\
Mini Batch Size                 & 32                   \\
PPO clip parameter                   & 0.2                 \\
GAE $\lambda$                   & 0.95                 \\
Discount Factor $\gamma$        & 0.99                 \\ \hline
\end{tabular}
}
\end{table}

\subsection{Modeling Real Sensors}

\begin{figure}[h]
    \centering
    \includegraphics[trim=0pt 0pt 0pt 0pt, clip, width=\linewidth]{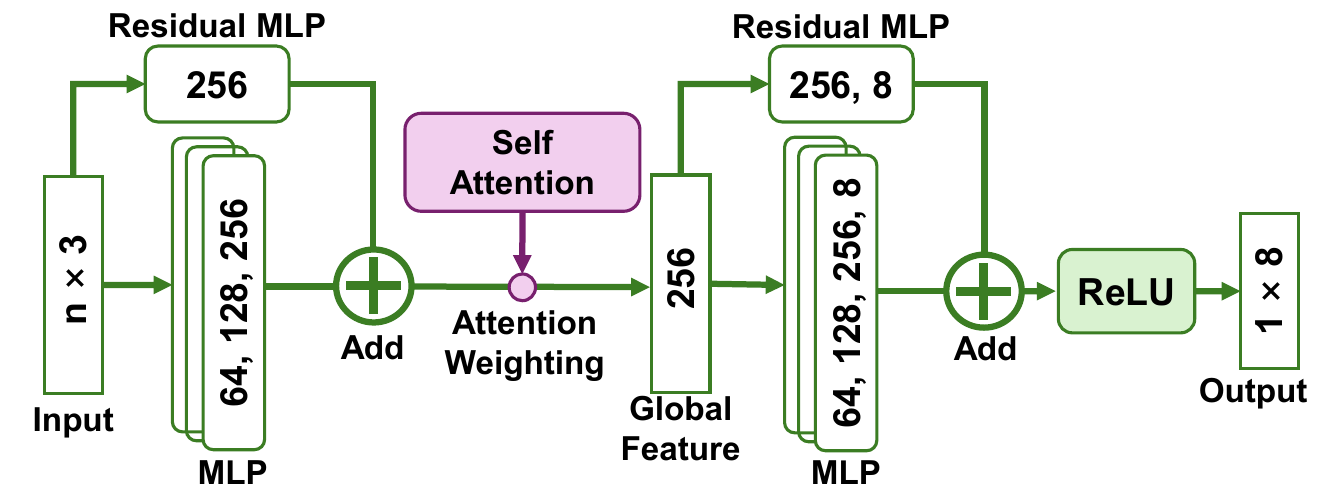}       \caption{Network used for sensor output prediction.}
    \label{fig:regressor}
\end{figure}

We trained a PointNet-based regressor to map simulated displacement fields to real sensor outputs (Sec.~\ref{sec: exp1}). The architecture of the regressor is illustrated in Fig.~\ref{fig:regressor}.  Residual connections are added to improve gradient back propagation, while attention weighting replaces max pooling in global feature extraction (following~\cite{yang2020attpnet}). 
The self-attention module consists of an MLP with shape [256, 128, 1], with tanh as activation. The resulting attention scores are normalized using a softmax function.

\bibliographystyle{ieeetr}
\bibliography{IEEEabrv}

\end{document}